\documentclass[times, twocolumn,final, nopreprintline]{elsarticle}
\usepackage{framed,multirow}

\usepackage{amssymb}
\usepackage{latexsym}

\usepackage{url}
\usepackage{xcolor}

\usepackage{hyperref}

\usepackage{amsmath, amsfonts}
\usepackage{algorithmic}
\usepackage{graphicx}
\usepackage{textcomp}
\usepackage{multirow}
\usepackage{makecell}
\usepackage{arydshln}
\usepackage{xr}
\usepackage{fancyhdr}

\fancypagestyle{allstyle}
{
   \fancyhf{}
   \fancyhead[L]{Manuscript accepted for publication in Medical Image Analysis (\href{https://doi.org/10.1016/j.media.2023.102940}{doi:10.1016/j.media.2023.102940})}
   \fancyfoot[C]{\thepage}
}
\fancypagestyle{firststyle}
{
   \fancyfoot[L]{\copyright 2023. This manuscript version is made available under the CC-BY-NC-ND 4.0 license \href{https://creativecommons.org/licenses/by-nc-nd/4.0/}{https://creativecommons.org/licenses/by-nc-nd/4.0/}}
}
\definecolor{newcolor}{rgb}{.8,.349,.1}

\journal{Medical Image Analysis}

\begin{document}

\begin{frontmatter}

\title{Deformation equivariant cross-modality image synthesis with paired non-aligned training data}%

\author[1]{Joel Honkamaa\corref{cor1}}
\cortext[cor1]{Corresponding author: Joel Honkamaa, PO Box 11000, FI-00076 Aalto, Finland}
\ead{joel.honkamaa@aalto.fi}
\author[2]{Umair Khan}
\author[3]{Sonja Koivukoski}
\author[4]{Mira Valkonen}
\author[3]{Leena Latonen}
\author[2]{Pekka Ruusuvuori}
\author[1]{Pekka Marttinen}

\address[1]{Department of Computer Science, Aalto University, Finland}
\address[2]{Institute of Biomedicine, University of Turku, Finland}
\address[3]{Institute of Biomedicine, University of Eastern Finland, Kuopio, Finland}
\address[4]{Faculty of Medicine and Health Technology, Tampere University, Finland}

\begin{abstract}
%%%
Cross-modality image synthesis is an active research topic with multiple medical clinically relevant applications. Recently, methods allowing training with paired but misaligned data have started to emerge. However, no robust and well-performing methods applicable to a wide range of real world data sets exist. In this work, we propose a generic solution to the problem of cross-modality image synthesis with paired but non-aligned data by introducing new deformation equivariance encouraging loss functions. The method consists of joint training of an image synthesis network together with separate registration networks and allows adversarial training conditioned on the input even with misaligned data. The work lowers the bar for new clinical applications by allowing effortless training of cross-modality image synthesis networks for more difficult data sets.
%%%%
\end{abstract}


\end{frontmatter}
\thispagestyle{firststyle}
\section{Introduction}
\label{sec:introduction}
Image-to-image translation is an active area of research in computer vision because of its various applications such as image synthesis, segmentation, restoration, style transformation and pose estimation. After the advent of deep leaning, medical imaging as a cardinal application area, has seen an increasing interest in the use of image-to-image translation. In histopathology, image-to-image translation has been used, e.g., for cross-stain translation \citep{liu2021unpaired, xu2019gan}, for replacing chemical staining by digitally generated mask \citep{valkonen2019cytokeratin}, for tissue color normalization \citep{de2018stain, de2021residual}, for virtual staining of label-free or unstained tissue images \citep{bayramoglu2017towards, rana2020use, rivenson2019phasestain}. In radiology, it has been used for pseudo CT generation and cross-modality MRI synthesis, and the synthesized images have been shown to be useful for downstream tasks, e.g., segmentation \citep{boulanger2021deep, spadea2021deep, xie2022survey}. The image-to-image translation methods are primarily divided into two categories: supervised methods that rely upon aligned image pairs and unsupervised methods that don't require aligned image pairs. In the medical setting including supervision tends to improve the results \citep{jin2019deep, klages2020patch, li2020comparison, peng2020magnetic, fard2022cnns}.

Image-to-image translation is called differently depending on the application and in this work we will call it cross-modality image synthesis, which is often used in the medical imaging context. We use the term modality broadly to refer to any distinct image types capturing different characteristics of the underlying anatomy.

In the medical domain, different modality images of the same subject are not usually anatomically aligned. To solve this before training a network images are typically registered, or in other words, aligned anatomically. Deep learning registration methods have gained popularity \citep{fu2020deep} with the best methods performing close to classical registration algorithms, e.g. in Learn2Reg multi-task medical image registation challenge \citep{hering2021learn2reg} or in histopathology ANHIR competition \citep{borovec2020anhir}.

Methods combining the two, cross-modality image synthesis and cross-modality registration, have also started to surface. In registration a synthesized image can be used as a bridge to generate a cross-modality similarity metric \citep{lu2021image}. However, some methods combine these two into a unified architecture, solving both problems at the same time. Such methods have been published from both the registration \citep{arar2020unsupervised, chen2022unsupervised} and image synthesis viewpoint \citep{joyce2017robust, kong2021breaking, wang2018unsupervised, wang2019tpsdicyc, wang2021dicyc}.

\begin{figure}[t]
\centerline{\includegraphics[width=\columnwidth]{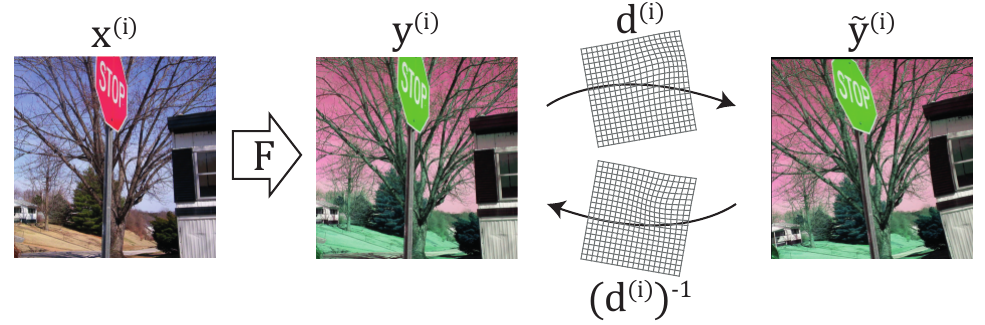}}
\caption{Basic setting. Only non-aligned pairs of $x^{(i)}$ and $\tilde{y}^{(i)}$ are available but the task is to learn $F$ for transforming $x^{(i)}$ into $y^{(i)}$. The images are from the synthetic "multimodal" data sets built using COCO \citep{lin2014microsoft} data set.}
\label{fig:basic_setting}
\end{figure}

In this paper, we propose a new architecture for cross-modality image synthesis which is robustly trainable with misaligned training data, using a novel strategy that couples registration and image synthesis networks during training. Firstly, we suggest training the image-synthesis network directly for deformation equivariance which refers to the property that applying a deformation before or after the image synthesis should result in the same image. Secondly, we develop a strategy allowing adversarial training conditioned on the input images despite using misaligned training data, not possible to do robustly by earlier methods. Conditioning adversarial training on the input is especially important in the medical domain as it results in more reliable predictions. In addition to the better quality predictions the method is applicable to a wider range of data sets than earlier similar methods. In the experiments we show that the method improves upon earlier methods trainable on non-aligned data and that it also surpasses the standard approach where the images are registered before training, assuming that no significant manual effort is put into the registration.

\section{Basic Setting}\label{sec:basic_setting}
Assume we have a paired training set of input images $(x^{(1)}, \dots, x^{(N)})$ and non-aligned target images $(\tilde{y}^{(1)}, \dots, \tilde{y}^{(N)})$. Following the notation by \citet{kong2021breaking} we denote the (unavailable) aligned ground truth targets by $(y^{(1)}, \dots, y^{(N)})$. Furthermore, unknown deformations $(d^{(1)}, \dots, d^{(N)})$ connect the coordinate systems of the inputs and the targets.

Assuming that the images are continuous, they can be seen as mappings $x^{(i)}: \mathbb{R}^n\to\mathbb{R}^{m_1}$ and $y^{(i)}, \tilde{y}^{(i)}: \mathbb{R}^n\to\mathbb{R}^{m_2}$ where $n$ is the dimensionality of the image (e.g. $n=2$ for two dimensional images) and $m_1$ and $m_2$ are number of channels in input and target images respectively (e.g. $3$ for RGB images). The deformations would then be mappings $d^{(i)}: \mathbb{R}^n\to\mathbb{R}^n$ connecting the image coordinates. Doing a coordinate transformation of an image $x^{(i)}$ based on a deformation $d^{(i)}$ would equal to function composition $x^{(i)} \circ d^{(i)}$ which can be written using the pullback notation as $d^{(i)*}x^{(i)} := x^{(i)} \circ d^{(i)}$ where $d^{(i)*}$ can be seen as a mapping acting on images.

Following the notation, we have the relationship $d^{(i)*}y^{(i)} = \tilde{y}^{(i)}$ between the aligned and non-aligned targets. In practice, the images are not continuous, but instead only samples of the images are available, i.e. the pixels or voxels. Hence in reality, the mapping $d^{(i)*}$ equals to interpolating the image at the locations defined by the deformation, and we use linear interpolation.

In this work, we study a setting where we are trying to learn a function $F$ which is a neural network such that $F(x^{(i)}) = y^{(i)}$. To do this, we simultaneously try to learn a second neural network, or, as it turns out, multiple networks, for predicting $d^{(i)}$. However, during test time we still only want to use the network $F$.

\section{Previous Work}
\subsection{Cross-Modality Image Synthesis}
Cross-modality medical image synthesis has gained a lot of attention in recent years with multiple proposed clinical applications \citep{wang2021review}. The conditional GAN-based architecture \textit{pix2pix} \citep{isola2017image} is widely used when paired and aligned data are available as it is based on an assumption of pixel-to-pixel correspondence between training images of different modality. On the other hand, \textit{CycleGAN} \citep{zhu2017unpaired} can be used without paired or aligned data.

Paired training images in the medical context are typically not aligned, and hence for pixel-to-pixel training they are registered into the same coordinate system. However, registration is never perfect and pixel-to-pixel losses are very sensitive to registration errors reducing the synthesis quality especially on difficult to register areas with large internal anatomic motion such as on pelvis area \citep{wang2021review}.

In pixel-to-pixel setting, different approaches have been proposed to mitigate for the remaining registration errors \citep{chen2020synthetic, joyce2017robust, kazemifar2019mri, leynes2018zero, yu2019ea}. Most similarly to our work, \citet{kong2021breaking} combine a cross-modality image synthesis network with a registration network to enable training with non-aligned data. However, we argue that their method does not robustly mitigate for registration errors especially with real world data sets as will be dicussed in Section \ref{sec:methods}. Additionally they use unconditional adversarial training which has been shown to be inferior to conditioning the discriminator with input images. In this work we aim to solve both of these problems.

While performing worse than \textit{pix2pix} when paired and aligned data are available, unsupervised \textit{CycleGAN} is more robust against misalignments due to its cycle consistency loss \citep{kaji2019overview, wang2021review}. However, if the misalignments between the modalities are systematic and severe, \textit{CycleGAN} can also fail to produce geometrically aligned predictions. Approaches, similar to ones used with \textit{pix2pix}, have also been employed with CycleGAN \citep{zhang2018translating, hiasa2018cross, kida2019cone}. \citet{wang2018unsupervised, wang2019tpsdicyc, wang2021dicyc} propose network architectures combining cross-modality image synthesis and registration, together with mutual information loss between the input and the prediction to promote similar geometry.

\subsection{Cross-Modality Registration}

Deformable medical image registration using deep learning has also gained popularity recently \citep{fu2020deep}. With stationary velocity field parametrization, one can generate diffeomorphic deformations \citep{arsigny2006log, ashburner2007fast}. This kind of methodology was applied to deep learning by \citet{dalca2018unsupervised}. Some architectures such as the one by \citet{de2019deep} combine affine or rigid registration together with a separate deformable registration resulting in multi-stage registration approach.

From cross-modality registration methods, the method by \citet{arar2020unsupervised} is closest to our work. They train a cross-modality registration network by simultaneously training a cross-modality image synthesis network which they encourage to be equivariant to deformations predicted by the registration network. This is done by applying the predicted deformation both before and after the image synthesis network and comparing both of them to the target. We instead use simulated deformations for encouraging deformation equivariance. We argue that this leads to a more robust approach, which we verify in the experiments. Our method of encouraging deformation equivariance is similar to the method by \citet{pielawski2020comir} where they train their network for rotational equivariance.

Very recently, \citet{chen2022unsupervised} use a contrastive learning based loss for promoting geometric (or shape) similarity of the image synthesis in an otherwise similar setting to Arar et al.

\section{Methods}\label{sec:methods}

\begin{figure}[!t]
\centerline{\includegraphics[width=\columnwidth]{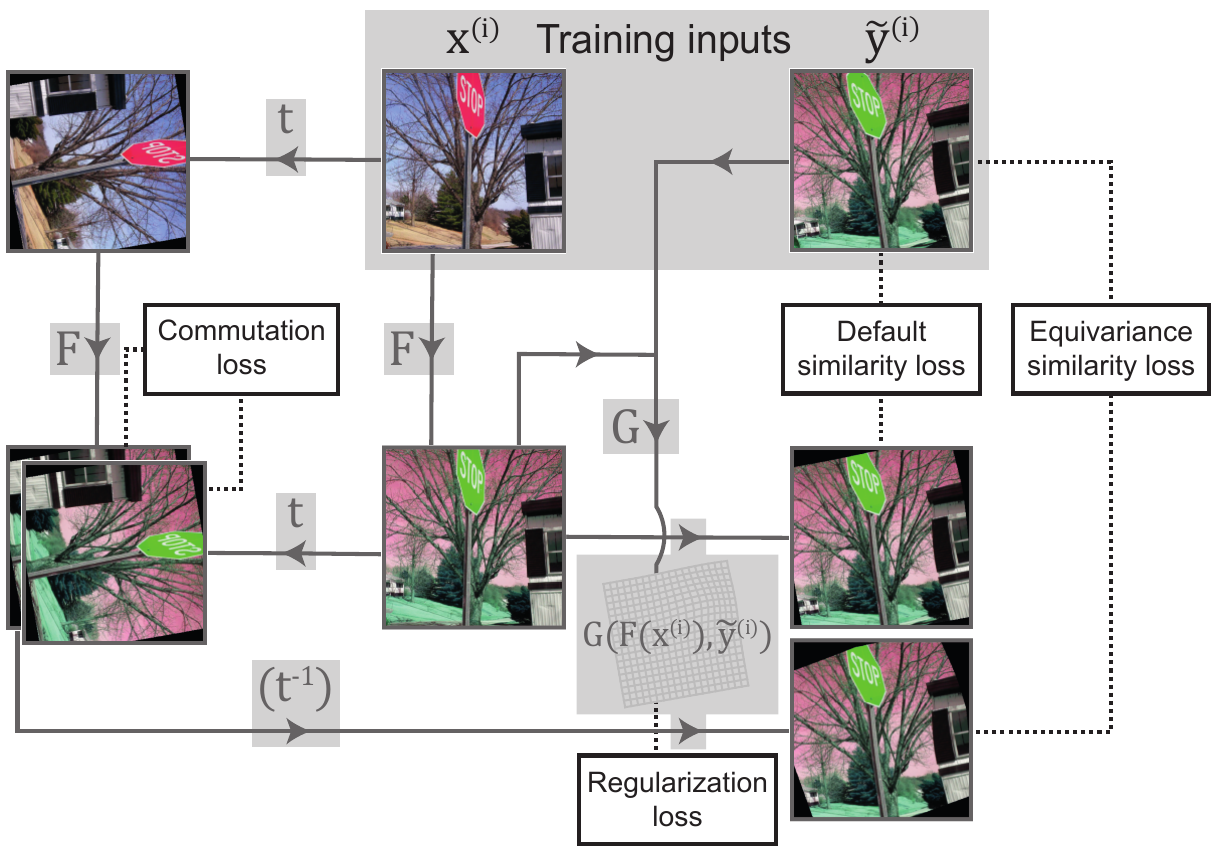}}
\caption{Proposed core architecture. When using the equivariance similarity loss instead of the default similarity loss, the commutation loss is optional. The architecture presented here is further refined by adding an adversarial loss and a separate cross-modality registration network for first registering $\tilde{y}^{(i)}$ to $x^{(i)}$. Deformation $t$ is a random deformation sampled on the fly individually for each training image pair. The images are from the synthetic "multimodal" data sets built using COCO \citep{lin2014microsoft} data set.}
\label{fig:core_architecture}
\end{figure}

To teach the network $F$ for predicting $y^{(i)}$ from $x^{(i)}$, \citet{kong2021breaking} train an additional network $G$ aimed at learning the $d^{(i)}$ s.t. $G(F(x^{(i)}), \tilde{y}^{(i)}) = d^{(i)}$. They train both of the networks $F$ and $G$ simultaneously with the similarity loss (which we label \textbf{default similarity loss})
\begin{equation}\label{eq:similarity_basic}
    \mathcal{L}_{\text{def-sim}} := \mathbb{E}_{x, \tilde{y}}||\tilde{y} - G(F(x), \tilde{y})^*F(x)||_{L^1}
\end{equation}
together with a \textbf{regularization loss}
$$\mathcal{L}_{\text{reg}} := \mathbb{E}_{x, \tilde{y}}\operatorname{Reg}(G(F(x), \tilde{y}))$$
where $\operatorname{Reg}$ is some operator penalizing non-smooth deformations, and an unconditional adversarial loss with the intent of training the distribution of $F(x^{(i)})$ to match the distribution of $\tilde{y}^{(i)}$.

In the work by Kong et al. they view the deformations between inputs and targets as noise and assume the same underling physical distribution for both the inputs and the targets. In that setting, the adversarial training objective they use is justified. However, often images in the target domain might be systematically geometrically different to the images in the input domain, e.g., when patients are laying differently within different medical imaging equipment. In that case matching the distribution of $F(x^{(i)})$ with the distribution of $\tilde{y}^{(i)}$ is not desirable.

To fix this we first omit the adversarial loss altogether, although we will develop a revised adversarial training strategy later. However, without the adversarial loss the optimization problem is very unstable since the network $F$ is not in any way constrained to preserve the geometry of the input images, that is, the predictions are not guaranteed to be anatomically aligned with the inputs. If $F$ is a convolutional neural network it has an inductive bias towards this kind of a behaviour but there is no guaranatee that the convolutional network does not, e.g., shift the predictions and the registration network compensate for the shift. An example of a possible failure mode is shown later in Figure \ref{fig:ixi_reggan_example}. It is also noteworthy that while having the adversarial training in a setting similar to the work by \citep{kong2021breaking} will definitely stabilize the training, there is no fundamental theoretical reason why it should result in $F$ preserving the geometry of its inputs.

The property of $F$ preserving the geometry of an input can be formulated as deformation equivariance. Any movement in the underlying anatomy of the input image should be reflected similarly in the output image. Assuming a set of anatomically possible geometric deformations $T^{(i)}$ for each input image $x^{(i)}$, a network $F$ is deformation equivariant over the deformations if for all $t \in T^{(i)}$ it holds that
\begin{equation}\label{eq:commutation}
t^*F(x^{(i)}) = F(t^*x^{(i)}).
\end{equation}
In other words, $F$ should commute with respect to the deformations of $x^{(i)}$.

In practice we do not aim for the relation to hold exactly but instead propose to promote the property implicitly by modifying the default similarity loss given in Equation \eqref{eq:similarity_basic}. The modification is similar to the one by \citet{pielawski2020comir}, although they use it in a different contrastive learning setting. We label the resulting loss \textbf{equivariance similarity loss}:
\begin{equation}\label{eq:similarity_implicit_eq}
    \mathcal{L}_{\text{eq-sim}} = \mathbb{E}_{x, \tilde{y}, t}||\tilde{y} - (G(F(x), \tilde{y})^*t^{-1})^*F(t^*x)||_{L^1}
\end{equation}
Here $t$ is seen as a random variable sampled from some distribution. The loss can be zero only if $F$ is equivariant to all $t$ and inputs. Note that we first compose the deformations $G(F(x), \tilde{y})$ and $t^{-1}$ and after that deform the prediction $F(t^*x)$. This way we avoid multiple interpolations of the same image. The same strategy of composing the deformations first is always used in this paper when applying multiple deformations to an image.

The loss in Equation \ref{eq:similarity_implicit_eq} also acts as a data augmentation method by randomly transforming inputs fed into the network $F$. However, in contrast to the traditional data augmentation, we do not transform the target image with the same deformation as the input.

Optionally one can more directly promote the equivariance by training with the following objective which we label \textbf{commutation loss}:
\begin{equation}\label{eq:commutation_loss}
\mathcal{L}_{\text{com}} := \mathbb{E}_{x, t}||t^*F(x) - F(t^*x)||_{L^1}
\end{equation}
When using the commutation loss using the equivariance similarity loss is optional and the default similarity loss can be used as well. Without the commutation loss the equivariance similarity loss is needed. As a result we have three possible configurations.

\citet{arar2020unsupervised} also encourage deformation equivariance but only for deformations predicted by their registration network. That is not always enough, e.g., with perfectly aligned training data the network $F$ could still introduce any translation which the registration network could compensate since translations commute. Also with subtle systematic deformations the network $F$ might easily overlearn the deformations from the data resulting in the registration network predicting zero deformation. The limitations of their method are also discussed in the supplementary materials of their paper where they conclude that their method works only with relatively small image synthesis networks, which is not a large problem in their work which focuses on image registration, but a significant limitation in difficult cross-modality image synthesis tasks. Later, in Figure \ref{fig:cermep_example}, we visualize a situation where our method succeeds in producing spatially aligned output but the method by \citet{arar2020unsupervised} (NeMAR) fails.

The core architecture presented so far is visualized in Figure \ref{fig:core_architecture}, and is in itself trainable. However, in addition to the core architecture, we will be looking at adding a conditional adversarial loss for training the model to improve the prediction quality further. In order for the adversarial training to converge even in the presence of systematically different geometries between the input and target domains, it turns out we will require two registration networks: one for registering targets to inputs and the other for registering predictions to possibly imperfectly registered targets.

Before advancing further, we introduce an additional notation. In case a variable should be treated as a constant from optimization point of view even if it is an output of a neural network, we overline the variable, e.g. $x^{(i)}$ vs. $\overline{x}^{(i)}$. In the neural network context, this means halting the backward pass during back-propagation.

\subsection{Selecting a Set of Simulated Deformations}\label{suplementary-sec:deformation_set}

The equivariance similarity loss and the commutation loss require some way to simulate anatomically realistic deformations for calculating them. For promoting equivariance, an ideal set would be the set of all anatomically realistic deformations for each sample, but anatomically realistic non-affine deformations are very difficult to simulate. A natural question is then whether a significantly smaller set of deformations would be sufficient, especially considering the inductive biases of the used architectures.

Given that $F$ is a convolutional network it is (roughly) translation equivariant. Hence intuitively simulating only globally affine deformations might be enough since any diffeomorphic deformation is "locally affine". Also, affine deformations are easy to sample and the same distribution of deformations can be used for each sample. Applying affine deformations to images is also computationally efficient and numerically accurate. For these reasons we used only affine transformations in our experiments. It is left for future work to test whether elastic deformations could increase the performance further.

Affine transformations can be further divided into translation, rotation, scaling, shearing, and flipping. Translations and rotations should be reasonable to use in any practical situations, and flipping can be used when it doesn't affect the distribution of the imaged anatomy. Scaling and shearing are more difficult since anatomically stretching a tissue could in principle affect its appearance under different imaging techniques. In other words, even in an ideal situation the deformation equivariance property would no longer hold exactly. In the experiments we conduct two studies on two data sets on using different types of affine transformations.

\subsection{Adversarial Training}\label{sec:adversarial}

In addition to the losses presented so far, we want to incorporate adversarial loss to the training in order to improve the appearance and also clinical quality of the predictions. In adversarial training, an additional discriminator network is trained to classify whether an image fed to the network is real or fake and can be used for guiding the generator network responsible for generating synthetic images. We want to employ a conditional adversarial training setting, similar to \textit{pix2pix} \citep{isola2017image}, wherein the input image is also fed to the discriminator. This is different to the approach taken by \citet{kong2021breaking} where they feed only the prediction or the target. Conditioning the discriminator on input images has potential for better prediction quality \citep{isola2017image}.

Let now $D$ be the discriminator network receiving an input image as the first argument and either a target or a prediction as the second argument. The conditional adversarial learning objective is defined as follows:
\begin{equation}\label{eq:placeholder_adversarial}
\begin{aligned}
    \mathbb{E}_{x, \tilde{y}}
[
    &\log D(\text{input}, \text{target})\\
    + &\log (1 - D(\text{input}, \text{prediction}))
]
\end{aligned}
\end{equation}
The discriminator is trained to maximize the loss and the generator is trained to minimize it with the training executed in turns while holding the weights of the other network constant. Placeholder texts are used as we are yet to derive the optimal way to feed the data to the discriminator. Misaligned training data will require care in how that is done. To be more precise, the following three points need to be taken into account:
\begin{enumerate}
    \item Predictions and targets have to be fed in the same coordinate system since the input domain and the target domain might have systematic geometric differences which would encourage the predictions to be misaligned with the inputs.
    \item Inputs and their corresponding predictions can not be fed in the exactly same coordinate system since even if the predictions are registered to the targets or vice versa, the targets will not be exactly aligned with the inputs, especially in the beginning of the training. That would encourage misaligned predictions.
    \item Interpolation acts as a low-pass filter especially in areas where the image is stretched. As a result, predictions registered to targets can not be directly compared with the targets as the discriminator will learn to notice the missing high frequencies.
\end{enumerate}
Points 1) and 2) would suggest to feed targets and predictions in the target coordinates but 3) would suggest that we can not deform the predictions to the targets either. As a solution we propose to train two separate registration networks: one for cross-modality registration of targets to inputs and one for intra-modality registration of predictions to possibly imperfectly registered targets. The adversarial comparison can then be done between the registered targets and the predictions registered to the registered targets. The proposed approach will solve all the three problems mentioned above: 1) The comparison will be done in the same coordinate system. 2) If the target and the input are imperfectly aligned, the prediction can still be separately registered to the registered target hence removing the incentive for misaligned predictions. 3) The predictions registered to the registered targets can contain high frequency information to at least a similar extent as their counterparts, the registered targets, since as the training progresses the registration of the targets to the inputs should account for most of the registration movement.

Let us now denote the predicted cross-modality deformation (approximately) mapping coordinates of $\tilde{y}^{(i)}$ to $x^{(i)}$ as $d^{(i)}_{\text{cross}}$ and the predicted intra-modality deformation (approximately) mapping coordinates of $F(x^{(i)})$ to ${d^{(i)}_{\text{cross}}}^*\tilde{y}^{(i)}$ as $d^{(i)}_{\text{intra}}$. We will look at how these are obtained in Section \ref{sec:registration}.

From the adversarial loss perspective, we want to treat $d^{(i)}_{\text{cross}}$ as constant since the cross-modality registration would not benefit from the adversarial loss and might even result in unexpected optima. This approach corresponds to the normal GAN training where only the second term is used in updating the generator. The proposed adversarial loss is then
\begin{equation}\label{eq:basic_adversarial}
\begin{aligned}
    \mathbb{E}_{x, \tilde{y}}
[
    &\log D(x, \overline{d}^*_{\text{cross}}\tilde{y})\\
    + &\log (1 - D(\text{x}, d_{\text{intra}}^*F(x))
].
\end{aligned}
\end{equation}

The loss function can be improved even further by employing a similar idea to the equivariance similarity loss. To simultaneously prevent discriminator from over-fitting and implicitly promote equivariance to a set of deformations, we propose to further modify the loss to the following form (which we label \textbf{equivariance adversarial loss}):
\begin{equation}\label{eq:refined_adversarial}
\begin{aligned}
    \mathcal{L}_{\text{eq-adv}} := \mathbb{E}_{x, \tilde{y}, t}
[
    &\log D(t^*x, (t^*\overline{d}_{\text{cross}})^*\tilde{y})\\
    + &\log (1 - D(t^*x, (t^*d_{\text{intra}}^*t^{-1})^*F(tx))
]
\end{aligned}
\end{equation}
Here, the same deformations $t$ can be used which are used for the equivariance similarity loss and the commutation loss. The core idea is to augment the inputs to the discriminator with the deformations $t$ while taking into account the unaligned nature of the training data.

\subsection{Registration Architecture}\label{sec:registration}

As discussed, the registration will be divided into cross-modality registration for registering targets to inputs and intra-modality registration for registering predictions to the registered targets. While the cross-modality registration receives pairs of different modality as inputs, it is trained with intra-modality loss based on the synthesised image $F(x^{(i)})$ similarily to the intra-modality registration.

In principle, the registration networks can predict the deformation in any suitable form. We split the cross-modality registration into rigid registration and elastic registration. The two-stage architecture makes it significantly easier for the model to handle large deformations. For intra-modality registration, we do not use two-stage architecture as cross-modality registration should take care of most of the registration movement.

We generate elastic deformations from stationary velocity fields to promote diffeomorphic deformations and to allow inverting the deformations. From a stationary velocity field the final diffeomorphic deformation is obtained by integrating the field over itself over a unit time. In group theory, this can be seen as exponentiation of a member of a lie algebra \citep{arsigny2006log}, and hence we denote the integration by $\exp$. Exponentiation can be estimated efficiently by the scaling and squaring method \citep{arsigny2006log, dalca2018unsupervised}. The velocity fields are predicted in the same resolution as the images.

\subsubsection{Cross-Modality Registration}

Let the neural network predicting the rigid deformation for cross-modality registration be $H_{\text{rig}}$ and the neural network predicting the stationary velocity field for elastic cross-modality registration be $H_{\text{svf}}$. Both networks $H_{\text{rig}}$ and $H_{\text{svf}}$ could be trained in principle with a single loss. However, it is possible that the rigid registration network could first unnecessarily shift the target image and the elastic registration network could then shift the image back, and to prevent this we add a separate rigid registration loss.

The overall predicted cross-modality deformation is then
$$
d^{(i)}_{\text{cross}} := \exp(v^{(i)}_{\text{cross}})^*\overline{r}^{(i)}_{\text{cross}}
$$
where $r^{(i)}_{\text{cross}} := H_{\text{rig}}(x^{(i)}, \tilde{y}^{(i)})$ and $v^{(i)}_{\text{cross}} := H_{\text{svf}}(x^{(i)}, \tilde{y}^{(i)})$. Halting the gradients for $r^{(i)}_{\text{cross}}$ is not necessary but makes loss function balancing more straightforward by separating the rigid registration altogether.

We train the rigid registration network $H_{\text{rig}}$ with the loss
\begin{equation}\label{eq:rigid_cross_modal_sim_loss}
\mathcal{L}_{\text{rig-sim}}^{\text{cross}} := \mathbb{E}_{x, \tilde{y}}[||\overline{F(x)} - r^*_{\text{cross}}\tilde{y}||_{L^1}]
\end{equation}
and the elastic registration network $H_{\text{svf}}$ with the loss
\begin{equation}\label{eq:elastic_sim_loss}
\begin{aligned}
\mathcal{L}_{\text{sim}}^{\text{cross}} :=
\mathbb{E}_{x, \tilde{y}}
[
&||\overline{F(x)} - d^*_{\text{cross}}\tilde{y}||_{L^1}
].
\end{aligned}
\end{equation}
We halt the gradients for the backward pass for $F(x)$ as we do not want the imperfect cross-modality and especially rigid cross-modality registered target image to affect the image synthesis network.

Additionally we need to regularize the deformation. The regularization term can be applied only to the elastic component as we do not penalize rigid deformations. We use non-rigidity penalty by \citet{staring2007rigidity} and apply it to both inverse and forward deformations. We have
\begin{equation}
\mathcal{L}_{\text{reg}}^{\text{cross}} := \mathbb{E}_{x, \tilde{y}}\left[\operatorname{Rig}(\exp(v_{\text{cross}})) + \operatorname{Rig}(\exp(-v_{\text{cross}}))\right]
\end{equation}
where $\operatorname{Rig}$ is the non-rigidity penalty by Staring et al. Details of the regularization used can be found in the supplementary materials.

\begin{figure}[t]
\centerline{\includegraphics[width=\columnwidth]{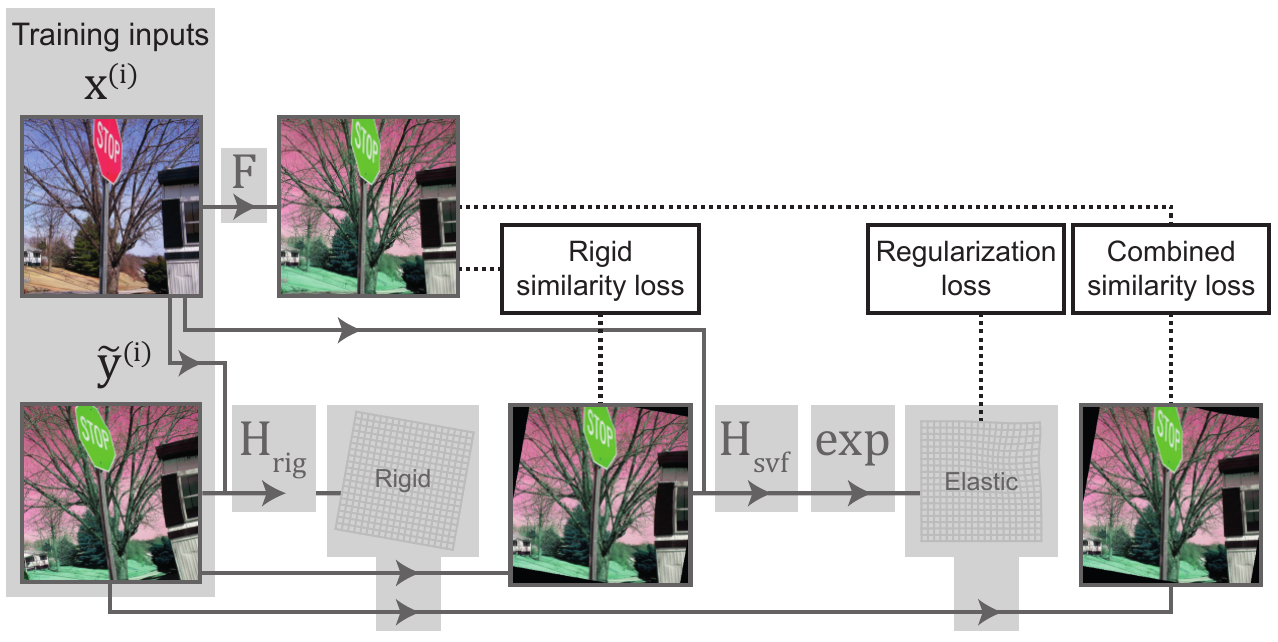}}
\caption{Cross-modality registration architecture. The outputs $y^{(i)}_{\text{reg}}$ and $v^{(i)}_{\text{cross}}$ are forwarded for intra-modality registration. Regularization is applied to both inverse and forward elastic deformation which is not explicitly shown here. The images are from the synthetic "multimodal" data sets built using COCO \citep{lin2014microsoft} data set.}
\label{fig:cross-modality_registration_architecture}
\end{figure}

\subsubsection{Intra-Modality Registration}

The intra-modality registration network receives the triplets $$(F(x^{(i)}), \overline{y}^{(i)}_{\text{reg}}, \exp(-\overline{v}^{(i)}_{\text{cross}}) - I)$$ as inputs where $I$ is the identity mapping. The third argument represents displacement field of the inverse elastic deformation with which $y^{(i)}_{\text{reg}}$ has been deformed and allows the network to optimize regularity of the concatenated overall deformation, as we regularize based on that.

Outputs of the cross-modality registration stage are treated as constants by the intra-modality registration losses. By that we prevent the networks from finding any non-desired optima where, e.g., difficult to synthesize regions were made smaller.

Let now the function predicting the stationary velocity field for intra-modality registration be $G_{\text{svf}}$. Then, the predicted intra-modality deformation is
$$
d^{(i)}_{\text{intra}} := \exp(-v^{(i)}_{\text{intra}})
$$
where $v^{(i)}_{\text{intra}} := G_{\text{svf}}(F(x^{(i)}), \overline{y}^{(i)}_{\text{reg}}, \exp(-\overline{v}^{(i)}_{\text{cross}}) - I)$. Here, we use the negative sign for the velocity field to emphasize that the direction is different from the cross-modality registration. As the training progresses and cross-modality registration and cross-modality image synthesis improve, $d^{(i)}_{\text{intra}}$ should approach the identity mapping.

The loss function for the intra-modality registration is also guiding the cross-modality image synthesis. Hence, we use the deformation equivarince encouraging loss function following the equation \eqref{eq:similarity_implicit_eq}:
\begin{equation}\label{eq:sim_loss_intra}
\begin{aligned}
\mathcal{L}_{\text{eq-sim}}^{\text{intra}}:=
\mathbb{E}_{x, \tilde{y}, t}
||(d^*_{\text{intra}}t^{-1})^*F(t^*x) - \overline{y}_{\text{reg}}||_{L^1}
\end{aligned}
\end{equation}

We also experiment with a setting where the $\mathcal{L}_{\text{sim}}^{\text{intra}}$ is replaced with the default similarity loss following the equation \eqref{eq:similarity_basic}. In that case, the loss simply takes the following form:
\begin{equation}\label{eq:sim_loss_intra_default}
\begin{aligned}
\mathcal{L}_{\text{def-sim}}^{\text{intra}} :=
\mathbb{E}_{x, \tilde{y}}
&||(d^*_{\text{intra}}F(x) - \overline{y}_{\text{reg}}||_{L^1}.
\end{aligned}
\end{equation}

For regularization, we use the concatenated overall elastic deformation again in both directions:
\begin{equation}
\begin{aligned}
\mathcal{L}_{\text{reg}}^{\text{intra}} := \frac{1}{2}\mathbb{E}_{x, \tilde{y}}[
&\operatorname{Rig}(\exp(v_{\text{intra}})^*\exp(\overline{v}_{\text{cross}}))\\ + &\operatorname{Rig}(\exp(-\overline{v}_{\text{cross}})^*\exp(-v_{\text{intra}}))
]
\end{aligned}
\end{equation}
Using the concatenated overall deformation is logical, as that is the deformation we are actually trying to learn and hence regularize.

Having the separate intra-modality registration network in addition to the cross-modality registration allows the prediction from the image synthesis network to directly affect the predicted deformation. As a result the cross-modality image synthesis network is efficiently optimized for generating predictions with lowest deformation regularization loss, which can be seen as a meaningful selection criteria among all the possible geometry preserving versions of the synthesized image.

\subsection{Masking}\label{suplementary-sec:masking}

Throughout the architecture, images are resampled by deformations, but the sampled locations might be outside the image. We connect each image with a mask that can initially represent invalid regions in the image. Each time an image is resampled, the mask is updated with the regions resampled from outside the image. In similarity losses, we then compare images only within intersections of the masks. Masks contain discrete values and no gradients flow through the masks during the backward-pass, preventing the optimization of the masks themselves. The same procedure is done also for the deformations as they are also resampled by other deformations.

Invalid region masks are also fed for the registration networks since for invalid regions only the regularization should affect the generated deformation.

Additionally, the masks of the registered targets and the predictions registered to the registered targets might be systematically different, which the discriminator could use for separating the images. We mitigate for that by multiplying each image fed to the discriminator by the intersection of the masks of the images compared.

\subsection{Overall Loss Function}

The overall loss function can be written as
\begin{equation}
\begin{aligned}
    \mathcal{L} &:=
    \overbrace{\mathcal{L}_{\text{rig-sim}}^{\text{cross}}}^{H_{\text{rig}}} +
    \overbrace{\mathcal{L}_{\text{sim}}^{\text{cross}} + \lambda \mathcal{L}_{\text{reg}}^{\text{cross}}}^{H_{\text{svf}}}\\
    &+ \underbrace{\mathcal{L}_{\text{sim}}^{\text{intra}} + \lambda \mathcal{L}_{\text{reg}}^{\text{intra}} + \gamma \mathcal{L}_{\text{com}} + \delta \overbrace{\mathcal{L}_{\text{eq-adv}}}^{D}}_{G_{\text{svf}}, F}
\end{aligned}
\end{equation}
where $\mathcal{L}_{\text{sim}}^{\text{intra}}$ is either $\mathcal{L}_{\text{eq-sim}}^{\text{intra}}$ or $\mathcal{L}_{\text{def-sim}}^{\text{intra}}$, and $\lambda, \gamma, \delta \in \mathbb{R}$ are loss function weights. Each loss component affects only the weights of the sub-networks written in the curly braces. Note that $D$ is trained to maximize the loss whereas the other networks are trained to minimize it. We use two optimizers, one for discriminator, and one for all the other components. This is different to the works by \citet{arar2020unsupervised} and \citet{kong2021breaking} which use a separate optimizer for the registration network.

Actual values used for the loss function weights are given in the supplementary materials.

\section{Experiments}\label{sec:experiments}

To evaluate our method, we conduct experiments on four diverse data sets of which two are real world medical imaging data sets, one is a semi-synthetic data set, and one is a synthetically constructed "multi-modal" data set. We perform an ablation study of the losses proposed and compare the method against multiple baselines. Additionally we evaluate on two data sets the performance when using different distributions of affine transformations with the commutation loss. The aim of the experiments is to establish to a reasonable extent:

\begin{enumerate}
    \item The performance of our method against earlier cross-modality image synthesis methods which are trainable on non-aligned data.
    \item The performance of our method against the standard pipeline where the image pairs are registered before training, assuming that no significant manual effort is put into registering the images.
    \item Which types of affine transformations are the most suitable for the equivariance encouraging losses and whether the choice has a large effect on the performance.
\end{enumerate}

On the two real world data sets we use clinically relevant metrics for establishing the best performance.

\subsection{Ablation Study}

We will use the following naming conventions to reflect loss terms to be included in different experimental configurations:
\begin{itemize}
    \item \textit{EqSim}: The equivariance similarity loss from Equation \eqref{eq:sim_loss_intra} was included.
    \item \textit{DefSim}: The default similarity loss from Equation \eqref{eq:sim_loss_intra_default} was included.
    \item \textit{Com}: The commutation loss from Equation \eqref{eq:commutation_loss} was included.
    \item \textit{EqAdv}: The equivariance adversarial loss from Equation \eqref{eq:refined_adversarial} was included.
    \item \textit{DefUncondAdv}: The default unconditional adversarial loss from \textit{pix2pix} \citep{isola2017image} defined directly between unmodified predictions and targets was included.
    \item \textit{NoReg}: Only the cross-modality image synthesis component $F$ with $L^1$ similarity loss directly between predictions and unmodified training targets was included.
    \item \textit{Aug}: Traditional data augmentation was used for each training input using the same distribution of deformations as what would have been used for the equivariance similarity loss, the commutation loss, and the equivariance adversarial loss.
\end{itemize}
The cross-modality registration related similarity loss and both of the regularization losses were used in all the trainings except with \textit{NoReg} setup.

In Section \ref{sec:methods} three variants of our developed method were proposed: \textit{EqSim}, \textit{DefSim} + \textit{Com}, and \textit{EqSim} + \textit{Com}. Optionally, \textit{EqAdv} can be combined with any of them. Training with any of the variants should result in a stable convergence and the experiments aim at measuring their relative performance.
\begin{table*}[t]
\centering
\caption{Deformation parameters for synthetic data sets}
\vspace{0.1in}
\setlength\tabcolsep{1pt}
\small
\begin{tabular}{c|cc|ccc}
\hline
&  \multicolumn{2}{c|}{Rigid}  & \multicolumn{3}{c}{Elastic}\\
 & Translation & Rotation & $\mu$ & $\sigma$ & $m$\\
\hline
LR & U($-15$, $15$) & U($-15^{\circ}$, $15^{\circ}$) & U($0$, $400$) & U($40$, $120$) & U($-20$, $20$)\\
SR & U($-1.5$, $1.5$) & U($-1.5^{\circ}$, $1.5^{\circ}$) & U($0$, $400$) & U($40$, $120$) & U($-2.0$, $2.0$)\\
LC & ($10$, $-10$) & $10^{\circ}$ & ($120$, $280$) & ($60$, $80$) & ($20$, $-20$)\\
SC & ($1$, $-1$) & $-1^{\circ}$ & ($120$, $280$) & ($60$, $80$) & ($2$, $-2$)\\
\hline
\multicolumn{6}{l}{$U$ refers to the uniform distribution independent for each dimension.}\\
\multicolumn{6}{l}{All the values except the rotations are in pixel coordinates.}\\
\end{tabular}
\label{table:synthetic_parameters}
\end{table*}
\subsection{Baselines}

We compare the method against four baselines:
\begin{itemize}
    \item \textit{Pix2pix} \citep{isola2017image}, trainable on paired aligned data.
    \item \textit{RegGAN} The method proposed by \citet{kong2021breaking}, trainable on paired unaligned data. We use the NICEGAN \citep{chen2020reusing} variant which uses $L^2$ adversarial loss as it performed the best.
    \item \textit{NeMAR} The method proposed by \citet{arar2020unsupervised}, trainable on paired unaligned data, originally suggested for image registration.
    \item \textit{CycleGAN} \citep{zhu2017unpaired}, unsupervised method trainable on unpaired data.
\end{itemize}

We use the official implementations \footnote{https://github.com/junyanz/pytorch-CycleGAN-and-pix2pix}\footnote{https://github.com/Kid-Liet/Reg-GAN}\footnote{https://github.com/moabarar/nemar} and modify them for our data sets.

For pix2pix and NeMAR we additionally train variants which use the same losses as the official implementations but our components and optimizers. We denote these variants by adding "our components" in parenthesis after the method name. For details, see Section S.IV of the supplementary materials. Note also that the method \textit{DefSim} + \textit{DefUncondAdv} + \textit{Aug} corresponds to training our architecture with the losses similar to RegGAN, although RegGAN uses a separate optimizer for the registration network.

Our proposed equivariance losses additionally act as data augmentation. To ensure that the reason for our methods performing better is not simply the effect of seeing more data, we augment the inputs for the baseline methods with the same distribution of deformations as is used for the equivariance similarity loss, the commutation loss, and the equivariance adversarial loss.

\subsection{Data sets}

\subsubsection{Synthetic}

We performed an ablation study on very simple synthetic "multimodal" data sets created using images from COCO data set \citep{lin2014microsoft} with unmodified images as input images. Target images were generated by circularly swapping the RGB color channels of the input images and by deforming them with simulated deformations. The simulated deformations were generated by a composition of rotation, translation, and an elastic deformation component generated by exponentiation of a stationary velocity field defined by parameters $\mu, \sigma, m \in \mathbb{R}^n$ using the formula
\begin{equation}\label{eq:simulated_elastic_component}
m_i \mathrm{e}^{-\frac{1}{2}\frac{||(x - \mu)||^2}{\sigma_i^2}},
\end{equation}
where $x$ is the spatial coordinate and $i \in {1, \dots, n}$ is the dimension ($n=2$ or $3$). Four data set were generated: LR (Large Random), SR (Small Random), LC (Large Constant), and SC (Small Constant). Used deformation parameters are displayed in Table \ref{table:synthetic_parameters}. We centrally cropped all the images to resolution $(400, 400)$, to avoid the need  to extrapolate values from outside the original image when synthetically deforming the images, for details see Section S.V of the supplementary materials. Training, validation, and test sets all contained $4113$ images.

With this data set, all the experiments were conducted without the adversarial loss to study separately the effects of deformation equivariance encouraging losses. The six models trained using each of the data sets are listed in Table \ref{table:results_synthetic}.

Compositions of the following transformations were used as simulated deformations for the loss functions and data augmentation:
\begin{enumerate}
    \item Rotations in range ($-15^{\circ}$, $15^{\circ}$)
    \item Orthogonal rotations of either $0^{\circ}$, $90^{\circ}$, $180^{\circ}$, or $270^{\circ}$,
    \item Random flips over any axis
\end{enumerate}
No model was trained with aligned data as it would be easily learned perfectly in this setup.

\subsubsection{Semi-Synthetic Cross-Modality Brain MRI Synthesis}

\begin{figure}[t]
\centerline{\includegraphics[width=\columnwidth]{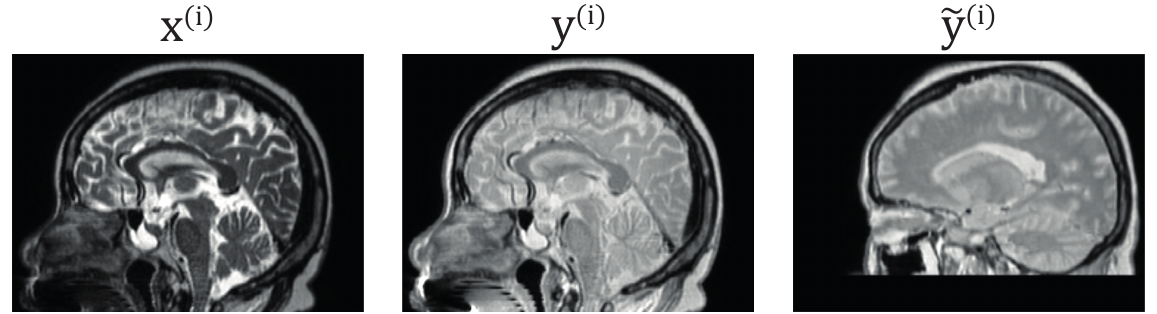}}
\caption{Example images from the semi-synthetic cross-modality MRI synthesis data set. Only one sagittal slide of the 3D volumes is visualized. The training target $\tilde{y}^{(i)}$ has been deformed with a random deformation after which we have cropped it such that the top and bottom edges are straight.}
\label{fig:ixi_training_data_example}
\end{figure}

In recent years, a significant amount of research has emerged on applying deep learning to cross-modality brain MRI synthesis. Synthetically generated modalities have many possible down-stream use cases such as segmentation, classification, detection and diagnosis. \citep{xie2022survey}

We used brain images from Information eXtraction from Images (IXI) data set \footnote{http://brain-development.org/ixi-data set/} to generate a semi-synthetic 3D data set for T2 to PD (proton density) synthesis, like in \citet{wang2021dicyc}. The main reason for choosing this task was that brain T2 and PD images are initially well aligned and hence provide good ground truths. Most of the images were already in resolution $(1.25\ \text{mm}, 0.9375\ \text{mm}, 0.9375\ \text{mm})$, the rest were also resampled to the same resolution. All the images were bias field corrected using N4 bias correction \citep{tustison2010n4itk} with Advanced Normalization Tools (ANTs) software \citep{avants2009advanced}, and normalized based on brain white-matter using the implementation by \citet{reinhold2019evaluating} together with the implementation by \citet{iglesias2011robust} for brain mask extraction by dividing all image values such that the white-matter had mean value of one in each image. We used $192$ images for training, $19$ for validation, and $365$ for testing.

To generate unaligned data set we applied simulated deformations to the target images. The deformations were generated by a composition of rotation, translation, and an elastic deformation component. Translations were sampled from range ($2.0\ \text{mm}$, $10.0\ \text{mm}$), rotations from range ($2.0^{\circ}$, $10.0^{\circ}$), and for elastic deformations we sampled white noise with mean of $10 \text{mm}$ and standard deviation of $200 \text{mm}$ followed by Gaussian smoothing with standard deviation of $10 \text{mm}$. The distribution is intentionally skewed to make the non-desired outcome of over-learning the deformation already in $F$ more attractive. We refer to the unaligned data set as "unaligned".

We re-register the synthetically deformed images using popular deformable registration method elastix \citep{klein2009elastix, shamonin2014fast} to compare our method with the standard approach of using registration as a pre-processing step. We refer to this data set as "registered".

Additionally we train an oracle model with the original aligned data set and refer to that data set as "aligned". For the oracle we use the same generator architecture as for our other methods. Its performance should provide a good upper bound on the performance of our methods.

For our methods we use 3D models and train them by sampling random image patches of size $(64, 64, 64)$ from the whole training data set. For 2D baseline models we used randomly sampled axial slices. Additionally the inputs were augmented with low-amplitude noise during the training. We also conducted a small experiment on the validation set for determining which type of simulated affine deformations are the most suitable ones for this problem. The experiment included translation, rotation, scaling and shearing. Flipping was not considered since the human anatomy is not symmetric.

\subsubsection{Virtual Histopathology Staining}

Virtual histopathology staining using deep learning has emerged as an active research topic in recent years, and has been primarily driven by GAN-based methods \citep{bayramoglu2017towards, rivenson2019phasestain, rana2020use, koivukoski2023unstained}. However, a majority of the methods require elastic registration of inputs and targets. Our method simplifies the data pre-processing by eliminating the need to elastically register image pairs explicitly.

We used a public data set containing unstained and stained tissue whole slides image (WSI) pairs \cite{khan2023effect} available at \footnote{https://doi.org/10.23729/9ddc2fc5-9bdb-404c-be07-c9c9540a32de}. These are essentially ultra high resolution gigapixel images, and virtually staining the unstained tissue WSIs is a highly non-trivial task. Pre-clinical murine prostate tissue samples were prepared at the University of Eastern Finland, Kuopio. Material used was surplus tissue from previous studies \citep{latonen2017vivo, valkonen2017analysis} where all animal experimentation and care procedures were carried out in accordance with guidelines and regulations of the national Animal Experiment Board of Finland, and were approved by the board of laboratory animal work of the State Provincial Offices of South Finland (licence number ESAVI/6271/04.10.03/2011). The tissue samples were first scanned without staining. This was followed by hematoxylin and eosin (H\&E) staining of the unstained tissue samples, and then the stained samples were scanned again. The samples were scanned using Thunder Imager 3D Tissue slide scanner (Leica Microsystems, Wetzlar, Germany) equipped with DMC2900 camera at 40X magnification level with a pixel size of 0.353µm. Total of 17 WSI pairs were included in the data set each with resolution of approximately $40\text{k} \times 40\text{k}$ from which $9$ were used for training, $1$ for validation, and $7$ for testing.

Inputs and targets were coarsely registered and the alignment seems superficially good. However, upon a closer inspection clear misalignments are present. We additionally registered the images using an open source cross-modality whole slide image registration tool called wsireg\footnote{https://github.com/NHPatterson/wsireg}. The WSI pairs were registered in two steps, first rigidly for global alignment and then elastically for more granular correspondence between the modalities. We refer to the coarsely registered original data set as "unaligned" and to the more finely registered data set as "registered". No oracle model was trained as we did not have ground truth registrations for this data set.

All of the models were trained by sampling random image patches of size $512 \times 512$ from the whole training data set. Additionally the inputs were augmented with low-amplitude noise during the training. On this data set we used the same set of transformations for the equivariance encouraging loss functions as was used in the synthetic experiment. We did not consider scaling or shearing for this data set as the misalignments on a single patch level are essentially rigid. Flipping was included as on the microscopic level the distribution should not be affected by that.

\subsubsection{Head MRI to CT synthesis}

\begin{figure}[t]
\centerline{\includegraphics[width=\columnwidth]{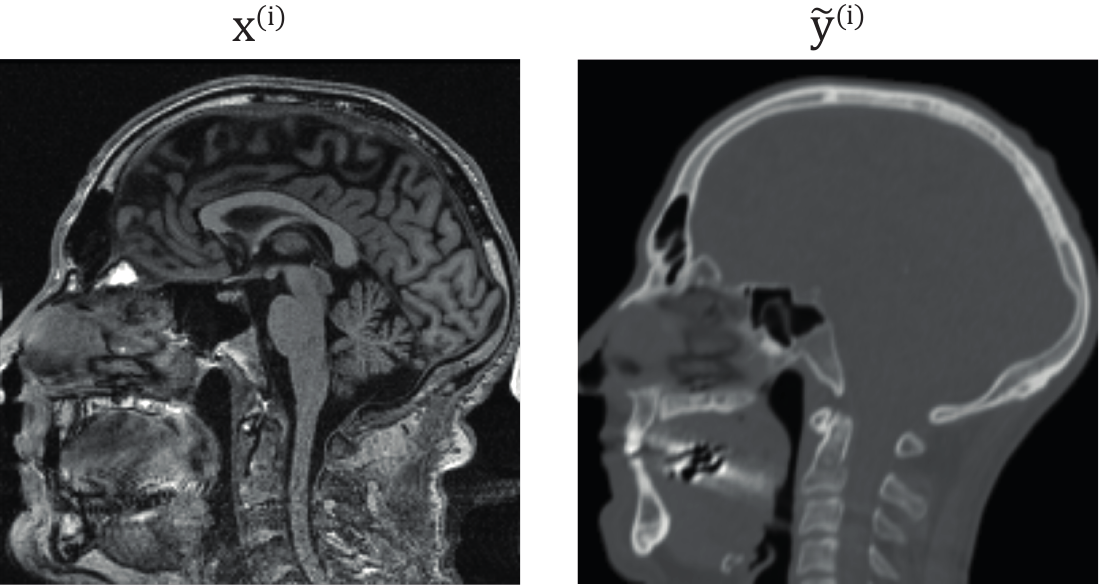}}
\caption{Example training images from the head MRI to CT synthesis data set. Only one sagittal slide of the 3D volumes is visualized. As can be seen, significant misalignments are present at the neck region. \copyright Copyright CERMEP – Imagerie du vivant, www.cermep.fr and Hospices Civils de Lyon. All rights reserved.}
\label{fig:cermep_training_data_example}
\end{figure}

Pseudo CT images are CT-like images generated from MRI images and are mostly used for replacing CT images in external beam radiation therapy (EBRT). Both predicting correct CT-values and geometrical accuracy of the generated images are important. In recent years Deep Learning has emerged as a strong option for pseudo CT generation.\citep{owrangi2018mri}

For the experiments we used CERMEP-IDB-MRXFDG data set which is freely available for research use \citep{merida2021cermep}. The data set consists of 37 rigidly registered CT and T1 MRI head scan pairs. We resampled all the images to the MRI-resolution of $160 \times 192 \times 192$. Pre-processing of the T1 images was similar to the one done for the cross-modality MRI synthesis data set as we applied N4 bias correction \citep{tustison2010n4itk} using Advanced Normalization Tools (ANTs) software \citep{avants2009advanced} and normalized the brain white-matter to the mean value of one using the implementation by \citet{reinhold2019evaluating} together with the implementation by \citet{iglesias2011robust} for brain mask extraction. We additionally removed any external objects from the CT images using series of morphological operations. While the skull and brain regions are relatively rigid, the image volumes extend to neck region with significant registration mismatches as can be seen in Figure \ref{fig:cermep_training_data_example}. We divided the data set to $20$ cases for training, $5$ for validation, and $12$ for testing.

We refer to the default data set as "unaligned". We additionally registered the images elastically using elastix \citep{klein2009elastix, shamonin2014fast} with the hyperparameters by \cite{leibfarth2013strategy} and refer to the data set as "registered".

Training setup was very similar to the that of cross-modality MRI synthesis experiment. We used 3D models and trained them by sampling random image patches of size $(64, 64, 64)$ from the whole training data set and for 2D baseline models we used randomly sampled axial slices. Additionally the inputs were augmented with low-amplitude noise during the training. We still similarly to the cross-modality MRI synthesis experiment conducted an experiment on the validation set for determining which type of simulated affine deformations are the most suitable. Flipping was again not considered since the human anatomy is not symmetric.

\subsection{Evaluation}

For all the experiments we measure structural similarity index (SSIM) \citep{wang2004image}, peak-signal-to-noise-ratio (PSNR), and normalized mutual information (NMI) \citep{studholme1999overlap}. NMI was applied between inputs and predictions as opposed to inputs and aligned targets as it is used here for measuring the geometric similarity of the predictions to the corresponding inputs. A detailed description of the pixel-wise metrics is given in the supplementary materials.

Additionally we measured the visual appearance of the synthesised images with Fréchet inception distance (FID) metric \cite{heusel2017gans, Seitzer2020FID}. While the visual appearance of the images is not usually clinically relevant, we still considered the comparison to be interesting enough to be included. For 3D data sets we computed FID over image slices over all three axes.

Evaluation of the virtual staining and pseudo CT experiments required more careful approaches described in the subsections \ref{sec:virtual_staining_evaluation} and \ref{sec:syn_ct_evaluation}.

\subsubsection{Virtual Histopathology Staining Evaluation}\label{sec:virtual_staining_evaluation}

We computed the pixel-wise metrics for the virtual staining data set separately for each predicted batch. As no ground truth was available, we registered affinely each stained patch to the corresponding unstained patch using Advanced Normalization Tools (ANTs) software \citep{avants2008symmetric, avants2009advanced}. We did not use directly the registered data set for computing the pixel-wise metrics to avoid the models trained with that data set from benefiting too much by being able to learn the exact registration dynamics (including possible systematic registration errors).

To further evaluate the quality of the virtually stained images, we conducted a comparative analysis of the virtual staining methods for nuclei reproducibility, a downstream validation approach similar to that of \citet{khan2023effect}. For that analysis we used the images from the registered data set as the ground truth. We employed the nuclei detection method by \citet{valkonen2020generalized} to output nuclei center coordinates in all the WSIs in the test set. First, nuclei were detected for the ground truth followed by nuclei detections in the virtually stained WSI generated by each of compared methods. F1-scores were computed to compare the detected nucleus coordinates of all outputs against those of the ground truth WSIs, using Euclidean distance with a tolerance of 5µm radius derived experimentally and through prior knowledge of typical nucleus dimensions \citep{lammerding2011mechanics, valkonen2020generalized}. We defined a true positive as a nucleus center in the virtually stained WSI for which there was a corresponding nucleus center in the ground truth WSI within 5µm radius. A false positive was defined as a nucleus center in the virtually stained WSI for which there was no corresponding nucleus center in the ground truth WSI within the 5µm radius. False negatives were nuclei centers in the ground truth WSI for which there were no matches in the virtually stained WSI within the 5µm radius.

\subsubsection{Head MRI to CT Synthesis Evaluation}\label{sec:syn_ct_evaluation}

For pseudo CT evaluation we additionally computed pixel-wise mean absolute error (MAE) and mean error (ME) as they are very widely used for pseudo CT evaluation and better predictors for resulting radiation dose differences than SSIM or PSNR \cite{boulanger2021deep}. ME refers to mean signed error over the data set, and can also be negative. ME largely ignores geometrical misalignements between inputs and predictions but on the other hand is robust to registration errors between inputs and targets used for evaluation.

We concluded the registered data set to be inadequate for accurate evaluation and had to develop more nuanced approach for registering the images, although still using the elastix software \citep{klein2009elastix, shamonin2014fast}. We noticed that the registration results improved by registering only part of the image at a time, probably since that way the rigid registration phase was able to account for a larger part of the total deformation. We ended up randomly sampling $20$ masks with radius of $10$ centimeters from each image and registered the image pairs over each of the masks. The evaluation metrics were computed over all the registrations with Gaussian weighting such that the highest weight was given to the coordinates at the center of the registration mask. Additionally we generated bone masks from the images by thresholding and applied the non-rigidity penalty \citep{staring2007rigidity} over those regions, improving the bone registration. That allowed us to use lower regularization value for the soft tissue regions improving the registration for those regions as well. We also manually masked out any regions with clear artefacts from the images. With these changes the quality of the registrations improved significantly based on visual evaluation. However, registration errors will always remain which will have to be taken into account in interpreting the results. To mitigate for registrations errors in body outline which end up easily dominating the metrics, we constructed body masks for both the MRI and CT images using morhoplogical operations and ignored in the evaluation the regions where the body masks did not match.

\section{Results and Discussion}

\subsection{Synthetic}

\begin{table}[t]
\centering
\caption{Results for the synthetic data set experiments}
\vspace{0.1in}
\setlength\tabcolsep{3pt}
\small
\begin{tabular}{clccc}
\hline
 Data set                  & Model            & PSNR   & SSIM      & NMI     \\
 \hline\multirow{6}{*}{LR} & EqSim + Com      & \textbf{36.45}  & \textbf{0.9842}    & \textbf{1.159}\\
                           & DefSim + Com & 33.95  & 0.9757    & 1.155\\
                           & EqSim            & 34.24  & 0.9781    & 1.154 \\
                           & DefSim           & -2.705 & 1.152e-03 & 1.042 \\
                           & DefSim + Aug     & -3.138 & 6.303e-04 & 1.049 \\
                           & NoReg + Aug   & 17.39  & 0.4550     & 1.079 \\
 \hline\multirow{6}{*}{SR} & EqSim + Com      & \textbf{41.68}  & \textbf{0.9954}    & \textbf{1.165}\\
                           & DefSim + Com     & 33.72  & 0.9737    & 1.153 \\
                           & EqSim            & 35.39  & 0.9817    & 1.155 \\
                           & DefSim           & 4.369  & 3.000e-03 & 1.022 \\
                           & DefSim + Aug     & 31.51  & 0.9559    & 1.147 \\
                           & NoReg + Aug   & 24.44  & 0.7607    & 1.123 \\
 \hline\multirow{6}{*}{LC} & EqSim + Com      & \textbf{38.78}  & \textbf{0.9905}    & \textbf{1.162} \\
                           & DefSim + Com     & 34.00     & 0.9745    & 1.153 \\
                           & EqSim            & 35.18  & 0.9819    & 1.155 \\
                           & DefSim           & 32.74  & 0.9640     & 1.149 \\
                           & DefSim + Aug     & -1.537 & 1.526e-03 & 1.039 \\
                           & NoReg + Aug   & 16.92  & 0.4617    & 1.073  \\
 \hline\multirow{6}{*}{SC} & EqSim + Com      & \textbf{40.89}  & 0.\textbf{9919}    & \textbf{1.166} \\
                           & DefSim + Com     & 34.05  & 0.9754    & 1.154 \\
                           & EqSim            & 35.89  & 0.9840     & 1.157 \\
                           & DefSim           & 15.93  & 0.38.00      & 1.070  \\
                           & DefSim + Aug     & 30.93  & 0.9501    & 1.145  \\
                           & NoReg + Aug   & 22.54  & 0.6696    & 1.114 \\
\hline
\end{tabular}
\label{table:results_synthetic}
\end{table}

\begin{figure}[t]
\centerline{\includegraphics[width=\columnwidth]{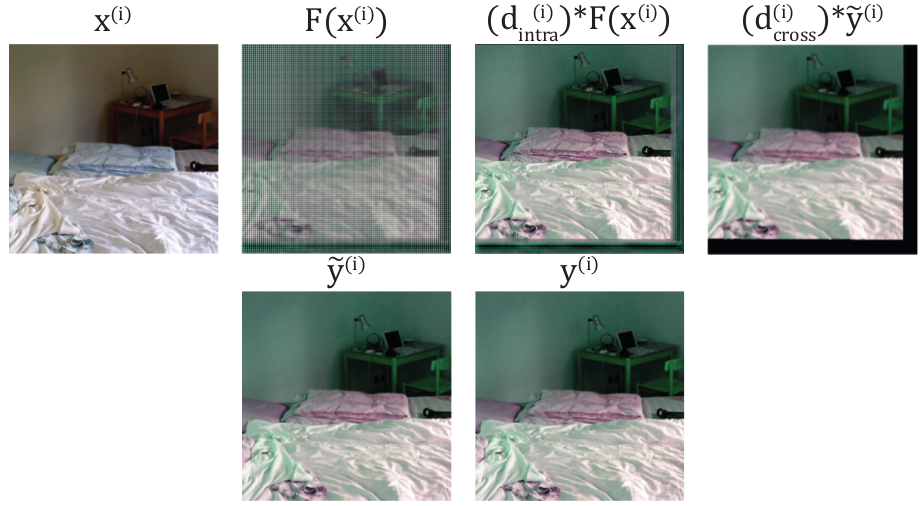}}
\caption{Example failure mode when training without deformation equivariance encouraging losses. The prediction is shifted towards top-left direction and also has a non-desired pattern both of which are compensated by the registration networks. Images are from the synthetic experiment with data set \textit{SR} and model \textit{DefSim}.}
\label{fig:synthetic_failure_example}
\end{figure}

\begin{figure}[t]
\centerline{\includegraphics[width=\columnwidth]{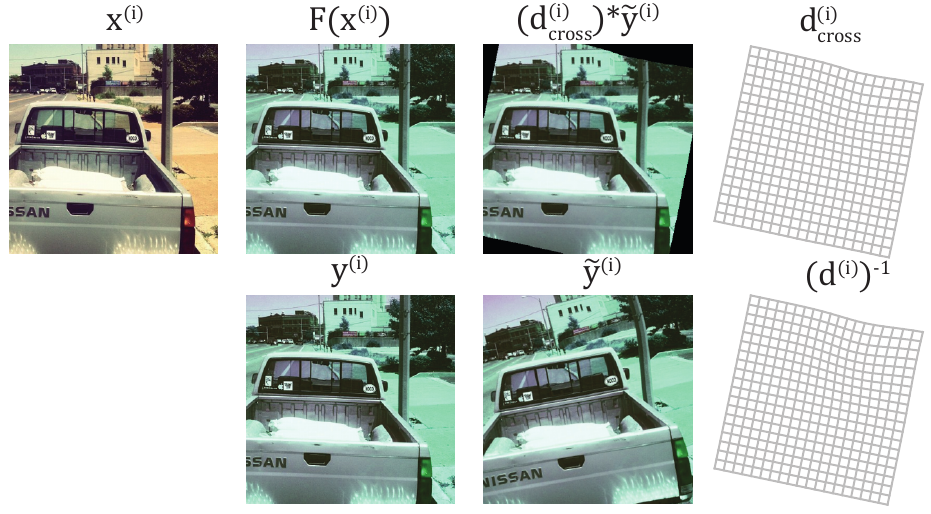}}
\caption{Example prediction from the synthetic experiment with data set \textit{LR} and model \textit{EqSim} + \textit{Com}. The image is from the test set. In addition to the synthesized image, the deformation is accurately reproduced.}
\label{fig:synthetic_example_prediction}
\end{figure}

Results for the synthetic data set experiment can be seen in Table \ref{table:results_synthetic}.

The models using the deformation equivariance encouraging losses systematically outperformed the models not using them. Four out of eight trainings with the registration component but without the deformation equivariance encouraging losses did not converge at all to a meaningful optimum. An example prediction of such a training is shown in Figure \ref{fig:synthetic_failure_example}. The performance of the models without the deformation equivariance losses varied a lot and in few cases the performance was even quite good. However, when using either of the deformation equivariance losses the trainings always converged robustly to a meaningful optimum.

Models using both the equivariance similarity loss and the commutation loss performed the best in terms of the similarity metrics. However, the models having only either the equivariance similarity loss or the commutation loss for encouraging deformation equivariance also performed very well and it is questionable whether the differences in performance when using this kind of synthetic data set will be relevant in real world applications.

\subsection{Cross-Modality Brain MRI synthesis}

\begin{table}[t]
\centering
\caption{Results for the cross-modality brain MRI synthesis experiment on the validation set comparing different types of affine transformations for the commutation loss. The experiment was performed using "DefSim + Com + EqAdv" setup. Translations were sampled from range $[-8\text{mm}, 8\text{mm}]$, and rotations from range $[-25^\circ, 25^\circ]$. Scales and shears were sampled by exponentiating a symmetric matrix with each matrix element being sampled from a zero mean Gaussian distribution with standard deviation of $0.08$. For generating scales non-diagonal values were set to zero.}
\vspace{0.1in}
\setlength\tabcolsep{1pt}
\small
\begin{tabular}{lccc}
\hline
 Transformations                             & PSNR  & SSIM   & NMI   \\
 \hline Translation                           & 34.33 & 0.9332 & 1.111 \\
 Translation + rotation                      & 36.43 & 0.9578 & 1.122 \\
 Translation + rotation + scaling            & 36.60 & 0.9621 & 1.122 \\
 Translation + rotation + scaling + shearing & \textbf{36.90} & \textbf{0.9623} & \textbf{1.125} \\
\hline
\end{tabular}
\label{table:cross-modality_mr_results_transformation_types}
\end{table}

\begin{table*}[t]
\centering
\caption{Results for the cross-modality brain MRI synthesis experiment}
\setlength\tabcolsep{2pt}
\small
\begin{tabular}{ll|ccc|ccc}
\hline
 Data set                                         & Model                         & PSNR  & SSIM   & NMI   & FID (sag) & FID (cor) & FID (ax) \\
 \hline\multirow{8}{*}{\makecell[l]{Non-aligned}} & EqSim + Com + EqAdv           & 36.30 & 0.9564 & 1.123 & 5.069     & 6.283     & 6.764    \\
                                                  & DefSim + Com + EqAdv          & \textbf{36.70} & 0.9595 & \textbf{1.124} & 6.302     & 7.290     & 7.709    \\
                                                  & EqSim + EqAdv                 & 36.73 & \textbf{0.9609} & 1.119 & 3.647     & 5.204     & \textbf{4.981}    \\
                                                  & DefSim + DefUncondAdv + Aug  & 26.67 & 0.6286 & 1.049 & 37.94     & 28.48     & 23.80    \\
                                                  & RegGAN                        & 21.24 & 0.3037 & 1.013 & 48.70     & 59.52     & 28.51    \\
                                                  & NeMAR                         & 19.98 & 0.4592 & 1.036 & 147.3     & 170.0     & 243.1    \\
                                                  & NeMAR (our components)   & 34.83 & 0.9377 & 1.107 & \textbf{3.461}     & \textbf{4.270}     & 5.243    \\
                                                  & CycleGAN                      & 23.32 & 0.4046 & 1.027 & 21.06     & 19.30     & 12.59    \\
 \hline\multirow{2}{*}{\makecell[l]{Registered}}  & Pix2pix                       & 28.96 & 0.8895 & 1.095 & 12.94     & 16.32     & 17.80    \\
                                                  & Pix2pix (our components) & 35.12 & 0.9386 & 1.107 & 10.01     & 10.95     & 10.29    \\
 \hline\hline\multirow{1}{*}{\makecell[l]{Aligned}}     & Pix2pix (our components)                        & 38.14 & 0.9679 & 1.111 & 4.093     & 5.462     & 6.514    \\
\hline
\end{tabular}
\label{table:cross-modality_mr_results}
\end{table*}

\begin{figure}[b]
\centerline{\includegraphics[width=\columnwidth]{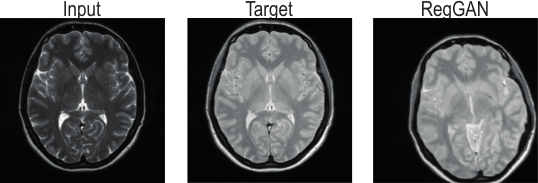}}
\caption{A prediction produced by the RegGAN model in the cross-modality brain MRI synthesis experiment. The model has converged to produce severely misaligned predictions.}
\label{fig:ixi_reggan_example}
\end{figure}

Results for the study on a validation set comparing different distributions of affine transformations for equivariance encouraging losses can be seen in Table \ref{table:cross-modality_mr_results_transformation_types}. Based on the study we used combination of all four transformation types in the main experiment for which the results with a test set can be seen in Table \ref{table:cross-modality_mr_results}.

All the three proposed variants of our method performed very well compared to the oracle model trained with aligned data and outperformed all the baselines with statistically significant margin on the voxel-wise metrics. NeMAR which also encourages deformation equivariance coupled with our 3D architecture is the only baseline trained on non-aligned data that came close to our method. RegGAN performed significantly worse and converged to produce severely misaligned predictions as visualized in Figure \ref{fig:ixi_reggan_example}. Pix2pix model trained on the registered data set coupled with our 3D architecture also performed well but was still clearly behind our methods while surpassing all the other models trained on non-aligned data. CycleGAN was also unable to converge to a meaningful optimum due to the large misalignments. While rarely directly relevant in clinical context, our method also performed very well in terms of the FID score.

\subsection{Virtual Histopathology Staining}

\begin{figure}[t]
\centerline{\includegraphics[width=\columnwidth]{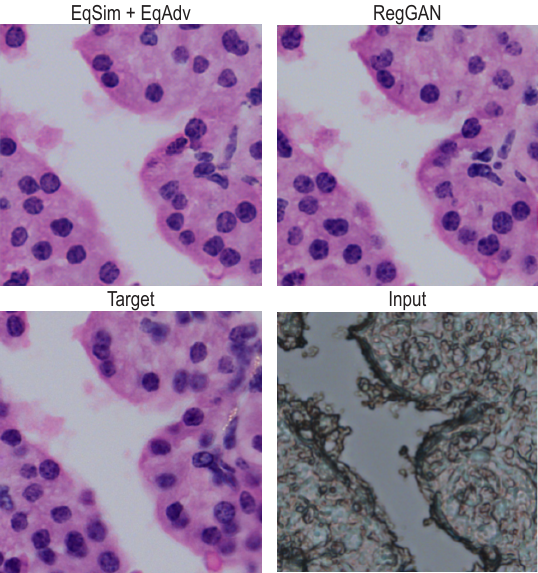}}
\caption{Virtually stained histopathological images with two top-performing models. An area of epithelial cells (pink) with nuclei (blue) is shown.}
\label{fig:virtual_staining_example}
\end{figure}

Results for the virtual histopathology staining experiment can be seen in Table \ref{table:results_virtual_staining}.

Nucleus reproduction from unstained brightfield images to virtual stained H\&E images is a particularly challenging task, as confirmed by the comparison of nucleus detection results between real and virtual stained images. This is in line with the conclusion from the earlier literature that the locations of all nuclei are simply not available in the unstained input images \citep{khan2023effect}. Additionally the data is not very uniform which also affects the evaluation metrics as training and test distributions do not match.

In terms of F1-score, which can be considered the main metric, RegGAN performed better than the other methods with statistical significance, and two of our methods were close behind, outperforming other baselines. We suspect that RegGAN benefited from having a generator with 24 times more parameters than our generator (1.1 billion vs. 46 million). Also, with this data set the distributions of the desired predictions $F(x^{(i)})$ and the training targets $\tilde{y}^{(i)}$ are very close to each other, i.e. there are no systematic geometrical differences. In such settings the RegGAN can be expected to perform well. It is left for future work to test our method with a larger generator.

Two of our configurations, \textit{EqSim} + \textit{EqAdv} and \textit{DefSim} + \textit{Com} + \textit{EqAdv}, beat both pix2pix variants trained on registered data in terms of F1-score by a statistically significant margin (p-values 0.016 and 0.00020). The result is more significant for the pix2pix model with our components as it was trained with identical architecture to our method. However, it is also noteworthy that the differences in loss function weightings affect the precision-recall balance which again can affect the F1-score, e.g. our training of the vanilla pix2pix had a different loss function balance affecting the precision-recall balance compared to the one trained with our components. For our proposed variants it seems that increasing deformation equivariance increases precision at the cost of recall. The model \textit{EqSim} + \textit{EqAdv} which did not have the commutation loss was more inclided to guess nuclei whereas the two other models with the commutation loss placed them in more certain locations. We suspect this is due to the commutation loss more directly promoting deformation equivariance which will require the shape of the nuclei to be known.

NeMAR adversarial training did not converge meaningfully with this data set due to the discriminator being easily able to distinguish between fake and real target images. We suspect that it was due to high frequency components present in this data set which the NeMAR architecture could not replicate for the discrminator due to the issues discussed in Section \ref{sec:adversarial}. Generator size of the unmodified NeMAR was also way too limited for the task.

\begin{table*}[t]
\centering
\caption{Results for the virtual histopathology staining experiment. Unmodified NeMAR was not included in the nuclei reproducibility study as it failed to converge to a meaningful optimum.}
\vspace{0.1in}
\setlength\tabcolsep{3pt}
\small
\begin{tabular}{ll|ccc|ccc|c}
\hline
&&\multicolumn{3}{c|}{Nuclei reproducibility}&\multicolumn{3}{c|}{}\\
 Data set                                         & Model                         & F1     & Precision & Recall & PSNR  & SSIM   & NMI   & FID   \\
 \hline\multirow{9}{*}{\makecell[l]{Non-aligned}} & EqSim + Com + EqAdv           & 0.7493 & 0.8514    & 0.6700 & 21.33 & 0.6491 & 1.020 & 43.04 \\
                                                  & DefSim + Com + EqAdv          & 0.7597 & 0.8291    & 0.7015 & 21.54 & 0.6553 & 1.019 & 35.00 \\
                                                  & EqSim + EqAdv                 & 0.7655 & 0.8004    & 0.7340 & 20.95 & 0.6358 & 1.018 & 21.51 \\
                                                  & DefSim + DefUncondAdv + Aug  & 0.7373 & 0.8838    & 0.6328 & 21.16 & 0.6409 & 1.019 & 42.70 \\
                                                  & RegGAN                        & \textbf{0.7799} & 0.7978    & \textbf{0.7647} & 21.22 & 0.6475 & 1.019 & \textbf{12.59} \\
                                                  & NeMAR                         & --     & --        & --     & 17.64 & 0.2791 & 1.014 & 333.1 \\
                                                  & NeMAR (our components)   & 0.6846 & 0.9151    & 0.5476 & 20.18 & 0.5788 & 1.015 & 55.35 \\
                                                  & CycleGAN                      & 0.5826 & 0.5712    & 0.5956 & 15.94 & 0.5073 & \textbf{1.039} & 31.50 \\
                                                  & Pix2pix (our components) & 0.7051 & 0.8762    & 0.5914 & 20.39 & 0.6144 & 1.016 & 40.82 \\
 \hline\multirow{2}{*}{\makecell[l]{Registered}}  & Pix2pix                       & 0.7068 & \textbf{0.9195}    & 0.5745 & \textbf{21.69} & \textbf{0.7025} & 1.023 & 51.73 \\
                                                  & Pix2pix (our components) & 0.7514 & 0.8628    & 0.6656 & 20.98 & 0.6705 & 1.020 & 30.28 \\
\hline
\end{tabular}
\label{table:results_virtual_staining}
\end{table*}

\subsection{Head MRI to CT Synthesis}

\begin{table*}
\centering
\caption{Results for the MRI to CT synthesis experiment on the validation set comparing different types of affine transformations for the commutation loss. The experiment was performed using "DefSim + Com + EqAdv" setup. Translations were sampled from range $[-8\text{mm}, 8\text{mm}]$, and rotations from range $[-25^\circ, 25^\circ]$. Scales and shears were sampled by exponentiating a symmetric matrix with each matrix element being sampled from a zero mean Gaussian distribution with standard deviation of $0.08$. For generating scales non-diagonal values were set to zero. Only translation and rotation were used for the main experiment as that resulted in the best MAE, although the difference is not very large in comparison to additionally using scaling and shearing. Note that ME largely ignores geometrical misalignements between inputs and predictions.}
\vspace{0.1in}
\setlength\tabcolsep{2pt}
\small
\begin{tabular}{lccccc}
\hline
 Transformations                             & MAE   & ME      & PSNR  & SSIM   & NMI   \\
 \hline Translation                           & 74.85 & \textbf{-0.2184} & 26.89 & 0.8699 & 1.067 \\
 Translation + rotation                      & \textbf{68.06} & -0.5071 & \textbf{27.50} & 0.8698 & 1.075 \\
 Translation + rotation + scaling            & 68.48 & 6.100   & 27.46 & 0.8723 & 1.077 \\
 Translation + rotation + scaling + shearing & 68.44 & 7.488   & 27.40 & \textbf{0.8727} & \textbf{1.077} \\
\hline
\end{tabular}
\label{table:pseudo_ct_results_transformation_types}
\end{table*}

\begin{table*}[t]
\centering
\caption{Results for MRI to CT synthesis experiment. Note that ME largely ignores geometrical misalignements between inputs and predictions.}
\vspace{0.1in}
\setlength\tabcolsep{3pt}
\small
\begin{tabular}{ll|ccccc|ccc}
\hline
 Data set                                         & Model                         & MAE   & ME      & PSNR  & SSIM   & NMI   & FID (sag) & FID (cor) & FID (ax) \\
 \hline\multirow{8}{*}{\makecell[l]{Non-aligned}} & EqSim + Com + EqAdv           & 67.59 & 9.091   & 28.41 & 0.8733 & \textbf{1.080} & 19.30     & 17.78     & 18.40    \\
                                                  & DefSim + Com + EqAdv          & \textbf{65.26} & 3.677   & 28.66 & 0.8794 & 1.078 & 22.66     & 20.33     & 20.79    \\
                                                  & EqSim + EqAdv                 & 66.08 & 1.752   & \textbf{28.82} & \textbf{0.8846} & 1.074 & 22.07     & 23.08     & 18.75    \\
                                                  & DefSim + DefUncondAdv + Aug  & 136.8 & \textbf{-0.8551} & 23.10 & 0.7431 & 1.049 & 49.51     & 31.95     & 34.90    \\
                                                  & RegGAN                        & 78.58 & -5.386  & 27.39 & 0.8461 & 1.069 & 63.80     & 49.06     & 18.47    \\
                                                  & NeMAR                         & 83.22 & -14.39  & 27.22 & 0.8331 & 1.071 & 57.13     & 46.16     & 36.28    \\
                                                  & NeMAR (our components)   & 67.49 & 11.60   & 28.57 & 0.8810 & 1.073 & \textbf{16.04}     & \textbf{15.14}     & \textbf{15.20}    \\
                                                  & CycleGAN                      & 99.56 & 8.469   & 25.73 & 0.8002 & 1.061 & 63.68     & 54.07     & 15.77    \\
 \hline\multirow{2}{*}{\makecell[l]{Registered}}  & Pix2pix                       & 100.9 & -3.347  & 25.82 & 0.7684 & 1.064 & 67.40     & 65.20     & 46.79    \\
                                                  & Pix2pix (our components) & 69.46 & -18.31  & 28.40 & 0.8787 & 1.071 & 28.64     & 25.11     & 22.16    \\
\hline
\end{tabular}
\label{table:results_pseudo_ct}
\end{table*}

\begin{figure}[t]
\centerline{\includegraphics[width=\columnwidth]{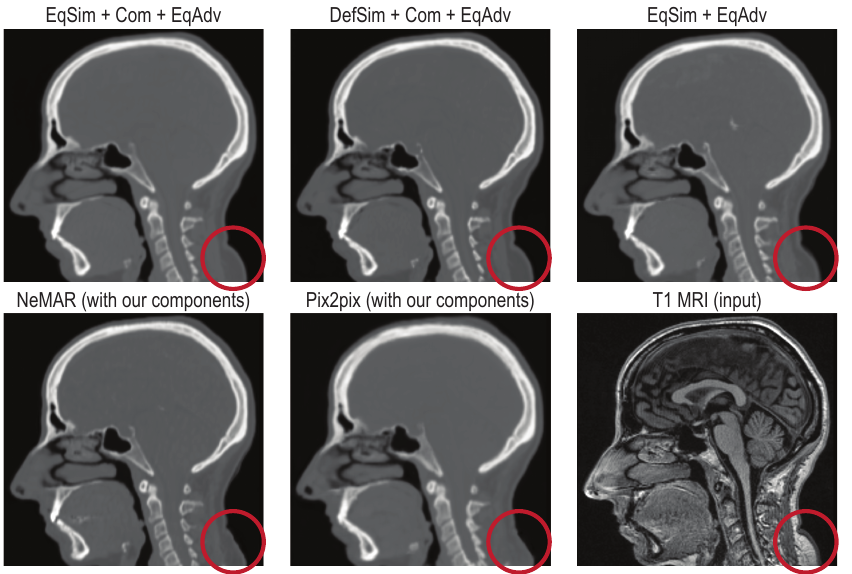}}
\caption{Predictions produced by different variants of our method and the two best performing baselines in the head MRI to CT synthesis experiment. Only our method is capable of generating the body outline at the lower neck region correctly. The region is highlighted with red circles. T1 MRI \copyright Copyright CERMEP – Imagerie du vivant, www.cermep.fr and Hospices Civils de Lyon. All rights reserved.}
\label{fig:cermep_example}
\end{figure}

Results for the study on validation set comparing different distributions of affine transformations for equivariance encouraging losses can be seen in Table \ref{table:pseudo_ct_results_transformation_types}. Based on the study we used only translation and rotation in the main experiment for which the results can be seen in Table \ref{table:results_pseudo_ct}. Note that the differences in metrics when using different affine transformation types are very small and might not be statistically significant.

Our proposed method performed very well in terms of the MAE which can be considered the most important metric for pseudo CT generation due to the strongly linear relationship between CT values and radiation absorption in radiation therapy. \textit{DefSim} + \textit{Com} + \textit{EqAdv} beat all of the baselines with statistically signifcant margin (p-value $0.0024$ compared to NeMAR with our components). \textit{EqSim} + \textit{EqAdv} also beat all of the baselines but when using the threshold of $0.05$ in p-value for statistical significance the difference in MAE is narrowly not significant (p-value $0.057$). Encouraging too much equivariance seems to be detrimental for correct CT-value estimation since \textit{EqSim} + \textit{Com} + \textit{EqAdv} performed slightly worse, although still not worse than any of the baselines.

While close to our method metric-wise, under visual inspection the images generated by NeMAR with our components contained more easily visible alignment mistakes than the images generated by our models. Probably the easiest mistake to notice was its inaccuracy in predicting the body outline at neck region, where the data set contains the largest systematic deformation differences. An example of such a case is shown in Figure \ref{fig:cermep_example}. More subtle mistakes included soft tissue boundaries being placed slightly off. Geometric accuracy of tissue boundaries is important as the pseudo CT images might also be used for positioning at the linear accelerator. The result is in line with the paper introducing NeMAR \citep{arar2020unsupervised} as in the supplementary materials they conclude that their image-to-image translation network produces geometrically accurate results only when the image synthesis generator model is significantly smaller than the one used here.

FID values have to be looked at with caution since they were calculated with respect to the unaligned data set whose distribution differs from that of the desired unavailable aligned CT images. However, based on visual inspection the NeMAR model with our components indeed produced the most realistic looking texture.

\section{Implementation}

Implementation of our method in PyTorch framework and all the evaluation implementations can be found at \href{https://github.com/honkamj/non-aligned-i2i}{https://github.com/honkamj/non-aligned-i2i}. The code base also contains all of the data pre-processing and allows for easily reproducing the results.

\section{Conclusions}

In this work, we have developed a generic method for training a network for cross-modality image synthesis with paired but misaligned training data by promoting equivariance with respect to simulated deformations. The method is applicable to a wider range of data sets than earlier methods and has the best overall performance accross three different cross-modality image synthesis tasks. On two tasks, cross-modality brain MRI synthesis and head MRI to CT synthesis, the method outperformed all of the baselines, and on the virtual staining task the performance was close to the best performing baseline, even though the baseline had a significantly larger network size. Based on the experiments while the \textit{EqSim} + \textit{Com} + \textit{EqAdv} configuration worked well on the synthetic data, our recommended configurations are \textit{EqSim} + \textit{EqAdv} and \textit{DefSim} + \textit{Com} + \textit{EqAdv} as they performed the best on more realistic data sets.

\section*{Acknowledgments}
This work was supported by Research Council of Finland (Flagship programme: Finnish Center for Artificial Intelligence [grant 345552] and grants 315896, 335976, 336033, 341967, 352986, 358246), ERA PerMed ABCAP (Research Council of Finland grants 334774, 334782), EU (H2020 grant 101016775 and NextGenerationEU). We also acknowledge the computational resources provided by the Aalto Science-IT Project.

%%Harvard
\bibliographystyle{model2-names.bst}\biboptions{authoryear}
\bibliography{refs}

\end{document}

\begin{frontmatter}

\title{Supplementary materials: Deformation equivariant cross-modality image synthesis with paired non-aligned training data}%

\author[1]{Joel Honkamaa}
\author[2]{Umair Khan}
\author[3]{Sonja Koivukoski}
\author[4]{Mira Valkonen}
\author[3]{Leena Latonen}
\author[2]{Pekka Ruusuvuori}
\author[1]{Pekka Marttinen}

\address[1]{Department of Computer Science, Aalto University, Finland}
\address[2]{Institute of Biomedicine, University of Turku, Finland}
\address[3]{Institute of Biomedicine, University of Eastern Finland, Kuopio, Finland}
\address[4]{Faculty of Medicine and Health Technology, Tampere University, Finland}

\end{frontmatter}

\thispagestyle{firststyle}
\section{Loss Function Weighting}\label{suplementary-sec:loss_weighting}

The following loss function weights were used for all the variations of our method:
\begin{itemize}
    \item The similarity loss terms were given the weight $1.0$.
    \item The commutation loss term was given the weight $1.0$.
    \item The adversarial loss terms were given the weight $0.0001$.
    \item The deformation regularization terms were given the weight $1.0$ in the cross-modality brain MRI synthesis and MRI to CT synthesis experiments and $100.0$ in the virtual staining experiment. In the synthetic COCO experiment we used the deformation regularization weight $0.1$ but the internal non-rigidity penalty weights were also different which is discussed in Section \ref{suplementary-sec:non-rigidity-penalty}.
\end{itemize}

\section{Training Details}\label{suplementary-sec:training_details}

Adam optimizer \citep{kingma2014adam} was used for training the models. In synthetic data set experiment learning rate $2e-4$ was used. In the other experiment learning rate $1e-4$ was used with a separate optimizer for the discriminator network with learning rate $4e-4$.

We used validation set for selecting the best epoch from last six training epochs for each model. Selection was done based on $L^1$ metric between ${d^{(i)}_{\text{intra}}}^*F(x^{(i)})$ and ${d^{(i)}_{\text{cross}}}^*\tilde{y}^{(i)}$. For models without the registration component $L^1$ metric between the prediction and the training label was used.

In experiments which used image patches for training, the patches were sampled randomly from the images. The training images contained invalid regions defined by masks and only patches containing a fully valid region in the input image were fed to the model. Our model can also handle partially invalid regions. However, because the baselines could not handle invalid regions, we used this training procedure for fair comparison.

\section{Network Architecture Details}\label{suplementary-sec:network_architectures}

For $F$, $H_{\text{svf}}$ and $G_{\text{svf}}$ \textit{U-net} \citep{ronneberger2015u} style convolutional networks with skip connections were used. Double convolution with kernel size $3$ was applied on each resolution followed by a downsampling layer or an upsampling layer. For downsampling we used strided convolutions with stride $2$ and kernel size $2$, and for upsampling corresponding transposed convolutions. For $F$, we used group normalization \citep{wu2018group} instead of typical batch normalization, while $H_{\text{svf}}$ and $G_{\text{svf}}$ had no normalization layers.

For $H_{\text{rig}}$ and $D$, \textit{ResNet} \citep{he2016deep} style encoders were used with global average pooling after a certain number of downsamplings followed by a linear mapping. For discriminator $D$ we additionally used spectral normalization \citep{miyato2018spectral} to make the training more stable. Also, we fed normalized image coordinates to the rigid registration network to handle rotations correctly despite of the global average pooling.

Leaky ReLU activation \citep{maas2013rectifier} with slope $0.01$ was used as non-linearity for $F$ and $D$ to avoid dead gradients as their optimization goal was suspected to change during the training. For all the other components ReLU activation was used.

Multiple inputs were concatenated over the channel dimension before feeding them to the networks.

In cross-modality brain MRI synthesis experiment and head MRI to CT synthesis experiments 3D architectures were used. Architectures used were otherwise similar to the architectures used in other experiments except that all 2D layers were replaced with corresponding 3D layers.

The following number of encoder features was used for convolutions of different components at each resolution for each experiment (for U-Net style networks this is followed by a decoder path with the number of convolution features in reverse order):

\begin{itemize}
    \item Synthetic
    \begin{itemize}
        \item $F$, $G_{\text{svf}}$, $H_{\text{rig}}$, and $H_{\text{svf}}$: $[32, 64, 128, 256, 256]$
    \end{itemize}
    \item Cross-Modality Brain MRI synthesis
    \begin{itemize}
        \item $F$, $D$: $[64, 128, 256, 512]$
        \item $G_{\text{svf}}$, $H_{\text{rig}}$, and $H_{\text{svf}}$: $[32, 64, 128, 256, 256]$
    \end{itemize}
    \item Virtual Histopathology Staining
    \begin{itemize}
        \item $F$: $[64, 128, 256, 512, 1024]$
        \item $D$: $[64, 128, 256, 512, 1024]$
        \item $G_{\text{svf}}$, $H_{\text{rig}}$, and $H_{\text{svf}}$: $[32, 64, 128, 256, 256]$
    \end{itemize}
    \item Head MRI to CT synthesis
    \begin{itemize}
        \item $F$, $D$: $[64, 128, 256, 512]$
        \item $G_{\text{svf}}$, $H_{\text{rig}}$, and $H_{\text{svf}}$: $[32, 64, 128, 256, 256]$
    \end{itemize}
\end{itemize}

\section{Baseline Network Architecture Details}

For two baselines, NeMAR \citep{arar2020unsupervised} and Pix2pix \citep{isola2017image}, we additionally trained variants for which the overall architecture was kept the same but subcomponents were replaced with the components used for our method. To be more specific, this included the following changes for any given model:
\begin{itemize}
    \item Generator and discriminator networks were replaced with our architectures described for each experiment in Section \ref{suplementary-sec:network_architectures} ($F$ and $D$).
    \item For NeMAR, the registration network was replaced with the architecture of $G_{\text{svf}}$ described in Section \ref{suplementary-sec:network_architectures} (together with the svf exponentiation).
    \item The optimizer described in Section \ref{suplementary-sec:training_details} was used.
    \item Loss function weights for each component were chosen to be identical with the loss function weights used for training our models. This included weights for the adversarial loss, the similarity loss, and for NeMAR also the deformation regularization loss.
\end{itemize}

\section{Synthetic Data Generation Details}

For two data sets we generated non-aligned target images by applying synthetic deformations. In such a process the values extrapolated from outside the image volume are usually set to zero. However, the resulting border could be used by the registration network for partially inferring the applied deformation, and with real world data sets no such border naturally exists. That would unnaturally benefit deep learning registration based methods such as our method. For COCO data set we avoid this problem mostly by centrally cropping the images to $400 \times 400$ resolution. However, for some COCO images and for cross-modality brain MRI synthesis images values were still extrapolated from outside the original images. For that reason we masked the synthetically deformed target images such that we computationally found the largest interior rectangle within the available image values and set everything else to zero.

\section{Evaluation Metrics}\label{suplementary-sec:evaluation_metrics}

The used evaluation metrics are explained here.

\subsection{Peak-Signal-to-Noise-Ratio (PSNR)}

PSNR is calculated as $$10 \log_{10} \frac{\text{MAX}^2}{\text{MSE}}$$ where $\text{MAX}$ is maximum possible image intensity and $\text{MSE}$ is the mean squared error between the images. For synthetic COCO dataset and virtual staining data set $\text{MAX} = 255$ and for cross-modality brain MRI synthesis data set we used maximum value over the whole data set since MRI images do not have a definite upper bound. Same value was used in all the experiments. 

\subsection{Structural Similarity Index (SSIM)}

SSIM \citep{wang2004image} is calculated over small windows and then averaged over the whole volume. We used rectangular window with side length of $7$ pixels or voxels. Formula for each window is $$\frac{(2\mu_1 \mu_2 + c_1)(2\sigma_{12} + c_2)}{(\mu_1^2 + \mu_2^2 + c_1)(\sigma_1^2 + \sigma_2^2 + c_2)}$$ where $\mu_1$ and $\mu_2$ are means, $\sigma_1$ and $\sigma_2$ are standard deviations of the images $y_1$ and $y_2$, respectively, and $\sigma_{12}$ is the covariance of the images. Constants $c_1$ and $c_2$ stabilize the division with weak denominator. Standard values $c_1 := (0.01 \times \text{MAX})^2$ and $c_2 := (0.03 \times \text{MAX})^2$ were used where $\text{MAX}$ is again the maximum possible image intensity.

\subsection{Normalized Mutual Information (NMI)}

NMI \citep{studholme1999overlap} is calculated as $$\frac{H(x) + H(y)}{H(x, y)}$$ where $x$ and $y$ are the compared images and $H$ is entropy. Each pixel or voxel pair at some location from the two images is assumed to be an independent sample from some distribution and the entropy is estimated using a joint histogram of the image values.

\subsection{Mean Absolute Error (MAE)}

Mean pixel or voxel-wise absolute error, always positive.

\subsection{Mean Error (ME)}

Mean pixel or voxel-wise error, can also be negative unlike the MAE. Does not measure spatial alignment well but on the other hand is robust to registration errors.

\subsection{FID}

Fréchet inception distance (FID) aims to measure perceived visual similarity of the images compared to some ground truth images measured as distributional difference in feature activations of Inception v3 network when fed the compared image sets. For details, see the paper by \citet{heusel2017gans}.

\section{Non-Rigidity Penalty}\label{suplementary-sec:non-rigidity-penalty}
Non-rigidity penalty by \citet{staring2007rigidity} consist of three separate terms: affinity term, orthogonality term, and properness term. Let now $d: \mathbb{R}^n\to\mathbb{R}^n$ be some deformation. Let $\frac{\partial d_j}{\partial x_i}$ be spatial derivative over $i^{\text{th}}$ dimension of $j^{\text{th}}$ component. $\frac{\partial d}{\partial x}$ is then a function mapping spatial coordinates to local Jacobian matrices.

Affinity term penalizes locally non-affine deformations, $\operatorname{Affinity}(d)=0$ if $d$ is everywhere affine. It is defined as
$$
\operatorname{Affinity}(d) := \frac{1}{V}\ \int_{\mathbb{R}^n} \sum_{i=1}^n \sum_{j=1}^n \sum_{k=1}^n \left[\frac{\partial^2 d_k}{\partial x_i \partial x_j}(x)\right]^2\ \mathrm{d}.
$$

Orthogonality term penalizes deformations which are not locally orthogonal,  $\operatorname{Orthogonality}(d)=0$ if $d$ is everywhere orthogonal. It is defined as
$$
\operatorname{Orthogonality}(d) := \frac{1}{V}\ \int_{\mathbb{R}^n} \left\lVert\frac{\partial d}{\partial x}{\frac{\partial d}{\partial x}}^T- I_n\right\rVert_F(x)\ \mathrm{d}.
$$
where $I_n$ is an identity matrix of size $n \times n$ and $||\cdot||_F$ is Frobenius norm.

Properness term penalizes deformations which apply local scaling, $\operatorname{Properness}(d)=0$ if $d$ does not apply local scaling anywhere. It is defined as
$$
\operatorname{Properness}(d) := \frac{1}{V}\ \int_{\mathbb{R}^n} \left(\det\left[\frac{\partial d}{\partial x}\right](x) - 1\right)\ \mathrm{d} x.
$$

In all the terms $V$ is a normalizing constant denoting the volume of the space considered. In real world discrete setting, the integrals are replaced by sums and $V$ is the number of spatial locations over which the sum is taken. Spatial derivatives can be approximated by differences between adjacent pixels. Note that if the resolution is not uniform that has to be taken into account in estimating the derivatives.

The final non-rigidity penalty consists of a weighted sum of the three terms. We used weight $1.0$ for the affinity term, weight $0.1$ for the properness term, and weight $0.01$ for the orthogonality term, except with the synthetic COCO data set we used the weight $1.0$ also for the orthogonality term.

\bibliographystyle{model2-names.bst}\biboptions{authoryear}
\bibliography{refs}